\pdfoutput=1
\documentclass[letterpaper,10pt,conference]{ieeeconf}

\IEEEoverridecommandlockouts
\AddToHook{package/hyperref/before}{%
  \expandafter\let\csname NAT@parse\endcsname\undefined
}
\AddToHook{package/hyperref/after}{%
  \global\let\cvprblueurl\url
}
\AddToHook{begindocument/before}{%
  \definecolor{cvprblue}{rgb}{0.21,0.49,0.74}%
  \hypersetup{%
    colorlinks=true,%
    linkcolor=cvprblue,%
    citecolor=cvprblue,%
    urlcolor=cvprblue%
  }%
  \global\let\url\cvprblueurl
}

\usepackage[utf8]{inputenc}
\usepackage[T1]{fontenc}

\usepackage{graphicx}
\usepackage{amsmath,amssymb,amsfonts}
\usepackage{mathtools}
\usepackage{nicefrac}
\usepackage{booktabs}
\usepackage{multirow}
\usepackage{array}
\usepackage{caption}
\usepackage{subcaption}
\usepackage{float}
\usepackage[table]{xcolor}
\usepackage{url}
\usepackage{tcolorbox}
\usepackage{bbm}
\tcbuselibrary{skins,breakable}

\usepackage{microtype}

\usepackage{cite}

\usepackage{xspace}

\usepackage{bbm}

\graphicspath{{figures/}{./}}
\definecolor{ours}{HTML}{2f6e68}
\definecolor{win}{HTML}{2f7d4f}

\definecolor{takeawayblue}{RGB}{53,90,150}
\definecolor{takeawayback}{RGB}{242,247,255}
\newtcolorbox{takeaway}{%
  enhanced,
  breakable,
  colback=takeawayback,
  colframe=takeawayblue,
  boxrule=0.8pt,
  arc=3pt,
  left=8pt,
  right=8pt,
  top=10pt,
  bottom=7pt,
  before skip=0.8em,
  after skip=0.8em,
  fontupper=\small,
  title=Takeaway,
  attach boxed title to top left={xshift=8pt,yshift=-7pt},
  boxed title style={
    colback=takeawayblue,
    colframe=takeawayblue,
    boxrule=0pt,
    arc=1pt,
    left=6pt,
    right=6pt,
    top=2pt,
    bottom=2pt
  },
  fonttitle=\bfseries\color{white}
}

\usepackage[hidelinks,bookmarks=false]{hyperref}
\hypersetup{pdfauthor={Kevin David Hayes, Alex Hanson, Allen Tu, Pranav Asthana, Matthias Zwicker, Tom Goldstein}, pdftitle={RIDE-Splat: Reflection Removal for In-the-Wild Driving Environments}}

\title{\LARGE \bf
RIDE-Splat: Reflection Removal for In-the-Wild Driving Environments
}

\newcommand{\authorspace}{\hspace{1.25em}}
\author{%
Kevin David Hayes \authorspace
Alex Hanson \authorspace
Allen Tu \authorspace
Pranav Asthana \authorspace
Matthias Zwicker \authorspace
Tom Goldstein\\[0.45em]
\textnormal{University of Maryland, College Park}\\[0.25em]
\url{https://ridesplat.github.io}
\vspace{-5mm}
}

\begin{document}

\twocolumn[{%
\renewcommand\twocolumn[1][]{#1}%
\maketitle
\begin{center}
\includegraphics[width=\linewidth]{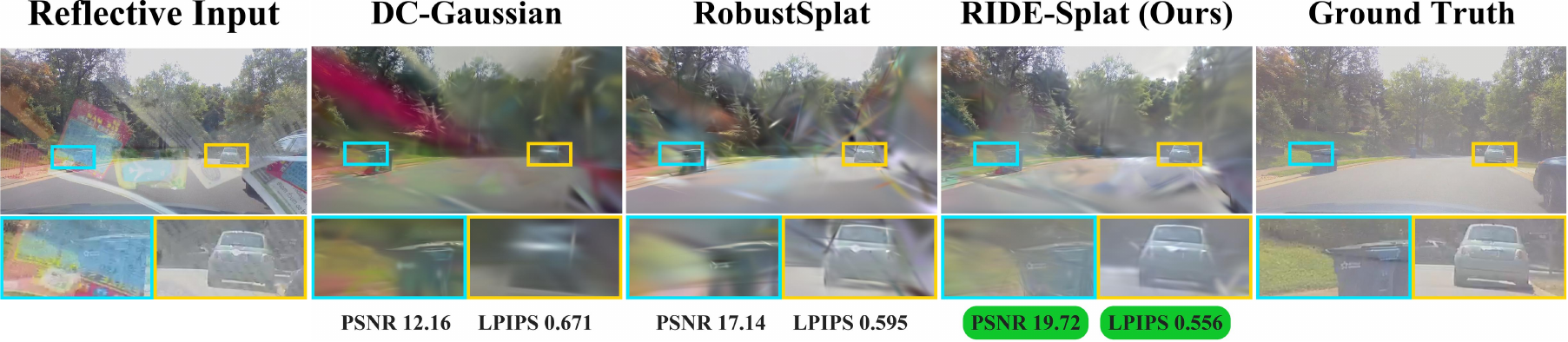}
\vspace{-4mm}
\captionsetup{type=figure}
\caption{\textbf{RIDE-Splat} reconstructs reflection-free novel views from driving video captured through a windshield. It separates the superimposed signals by representing the world-consistent transmission through the windshield with a 3D Gaussian Splatting scene~\cite{3dgs} and the dominant image-stationary windshield reflection with a shared, signed image. On the \emph{Reston} scene from our \textbf{RIDE-Splat Dataset}, the first and last columns show the nearest reflective training frame and the non-reflective ground-truth evaluation frame, respectively. RIDE-Splat removes reflections that persist in DC-Gaussian~\cite{dcgaussian} and RobustSplat~\cite{robustsplat} while preserving the transmitted scene.}
\label{fig:teaser}
\end{center}
}]

\begin{abstract}
Driving video captured through a windshield violates the world-space consistency assumed by 3D Gaussian Splatting (3DGS): the road scene exhibits parallax while cabin reflections remain attached to the camera, causing standard 3DGS to embed reflected content into the reconstructed scene. We introduce RIDE-Splat, a pipeline for reflection-free 3D reconstruction from posed driving video without clean images, reflection masks, or pretrained networks. RIDE-Splat assigns world-consistent transmission through the windshield to a 3DGS scene and the dominant camera-attached reflection to one shared, signed RGB image. It builds the clean reconstruction in two passes. First, it learns the 3D geometry and the signed reflection image, representing both added radiance and transmission attenuation. Then, it uses the reflection-removed training images to fit a clean 3DGS model. At inference, the reflection image is discarded, preserving the standard 3DGS rendering path. We also introduce the RIDE-Splat Dataset, the first 3D reconstruction benchmark for reflection removal from driving video -- with real driving scenes, physical windshield reflections in training views, and separately captured non-reflective evaluation images. Across the RIDE-Splat Dataset, XLD, and Wayve, RIDE-Splat outperforms 3DGS, DC-Gaussian, and RobustSplat on all nine aggregate metrics, improving PSNR by up to 3.59 dB and reducing LPIPS by up to 0.0876. We achieve the highest metrics in 38 out of 42 per-scene comparisons and removes windshield reflections that remain visible in the baselines while preserving the transmitted scene.
\end{abstract}

\section{Introduction}

3D Gaussian Splatting (3DGS) enables real-time, photorealistic novel-view synthesis~\cite{3dgs}, but its standard formulation assumes that posed images depict a coherent world-space scene. Driving video captured through a windshield violates this assumption because each frame contains the transmitted road scene and reflections of the vehicle interior. Camera motion reveals the key separation cue: the road scene exhibits parallax and remains consistent in world coordinates, while the dominant windshield reflection moves with the camera and remains approximately fixed in image coordinates. Standard 3DGS can therefore fit reflected content as scene geometry or appearance, contaminating novel views. Single-image reflection removal cannot exploit this multi-view cue, and rejecting contaminated observations discards valid transmission supervision. Conversely, a flexible per-view reflection field can mistakenly absorb road appearance that belongs in 3D. These failures motivate a constrained decomposition that assigns each signal to the coordinate system in which it remains consistent.

Historical dashcam archives and consumer driving videos document diverse roads, weather, lighting, and rare events that would be costly or impossible to recapture. Yet they remain underused for 3D reconstruction because windshield reflections make them poorly suited to pipelines designed for clean, purpose-captured imagery. Recovering reflection-free 3D scenes from these recordings could support closed-loop simulation, sensor-realistic augmentation, and rare-event replay without dedicated recapture.

We therefore introduce \textbf{RIDE-Splat}, a non-generative 3DGS formulation for reflection-free novel-view synthesis using only posed driving video with windshield reflections. It assigns world-consistent transmission through the windshield to a 3DGS scene and the dominant camera-attached reflection to one shared, signed RGB image. Because the reflection image is fixed in image coordinates and shared across views, it can represent image-stationary reflection content but cannot follow parallax-bearing road texture. Its signed values capture both added reflected radiance and transmission attenuation. The first pass uses a geometry warm-up to establish scene structure and a sparsity penalty to bias ambiguous content toward the 3D scene. Reflection self-localization then derives a fixed spatial weight from the recovered reflection image and refits both representations from scratch with the weight frozen. This second pass suppresses scene leakage without adding optimized parameters. At inference, we discard the reflection image, using standard 3DGS rasterization for rendering.

Existing benchmarks force a tradeoff: datasets with exact non-reflective targets use synthetic reflections, while those with physical windshield reflections lack clean targets. The \textbf{RIDE-Splat Dataset} closes this gap as the first 3D reconstruction benchmark designed specifically for reflection removal from driving video. For each real driving scene, it pairs a training traversal containing physical windshield reflections with a separately captured non-reflective evaluation traversal of the same route. The paired traversals are recorded within minutes using the same rigid camera mount, with only the reflection-inducing materials removed. All methods train only on the reflective traversal, with the non-reflective images held out as evaluation targets.

Across the RIDE-Splat Dataset, XLD~\cite{xld}, and Wayve~\cite{wayve}, RIDE-Splat outperforms 3DGS, DC-Gaussian~\cite{dcgaussian}, and RobustSplat~\cite{robustsplat} on every aggregate metric under both physical and synthetically added reflections. It improves PSNR by up to 3.59\,dB and LPIPS by up to 0.0876. These gains are consistent across scenes, with RIDE-Splat achieves higher performance in 38 of 42 per-scene metric comparisons. As shown in Fig.~\ref{fig:teaser}, RIDE-Splat removes windshield reflections that remain visible in the baselines while better preserving the transmitted scene.

In summary, we propose the following contributions:
\begin{enumerate}\itemsep2pt
\item \textbf{RIDE-Splat.} A coordinate-based 3DGS method that separates world-consistent transmission from camera-attached windshield reflections.
\item \textbf{RIDE-Splat Dataset.} The first 3D reconstruction benchmark for reflection removal from driving video.
\end{enumerate}

\section{Related Work}
Reflection removal has primarily been studied in image space.
Single-image methods separate transmission and reflection using handcrafted or learned priors~\cite{sir2benchmark}.
Multi-frame methods use layered motion, low-rank structure, or temporal consistency to distinguish the two signals~\cite{szeliski2000,nandoriya2017,liu2020obstructions}.
RIDE-Splat instead brings this motion cue into 3D reconstruction, assigning world-consistent transmission to the 3D scene and the dominant windshield reflection to image coordinates.

Neural radiance fields~\cite{nerf} and 3D Gaussian Splatting (3DGS)~\cite{3dgs} represent view-dependent radiance in world coordinates.
Reflection-aware variants of these representations model glossy appearance, separate transmitted and reflected radiance, or reconstruct specular and semi-transparent surfaces~\cite{refnerf,gaussianshader,refgaussian,nerfren,rtsplatting}.
Flash-Splat~\cite{flashsplat} separates transmission and reflection with Gaussian primitives but requires paired flash and no-flash captures.
RIDE-Splat instead targets driving videos and learns a windshield-reflection layer fixed in image coordinates from reflective monocular video alone.

In-the-wild 3D reconstruction methods suppress transient or view-inconsistent distractors through sample rejection or transient modeling~\cite{robustnerf,spotless,hybridgs}.
DF3DV-1K~\cite{lu2026df3dv} benchmarks anti-distractor 3DGS methods~\cite{t3dgs,wildgaussians,spotless,desplat,degauss,ocsplats,asymgs}, but does not include reflections of the vehicle interior on the windshield.
RobustSplat~\cite{robustsplat}, the strongest single-model baseline, delays Gaussian densification and bootstraps transient masks across scales.
Our reflective weight maps are inspired by SplatSuRe~\cite{splatsure} which derives per-pixel weight maps from multi-view sampling fidelity, but targets multi-view-consistent super-resolution rather than reflection removal.
Driving-scene methods reconstruct large environments and dynamic actors~\cite{streetgaussians,drivinggaussian,emernerf,neurad}, but do not model windshield reflections.

DC-Gaussian~\cite{dcgaussian} is the only prior 3D reconstruction method designed specifically to remove windshield reflections.
It predicts a per-view reflection field from camera pose using a hash grid and MLP, while a latent code modulates the field's intensity over time.
RIDE-Splat instead uses one shared, signed image to impose a coordinate-based separation prior; this representation is orthogonal to DC-Gaussian's geometry enhancement.
In Sec.~\ref{sec:results}, RIDE-Splat outperforms DC-Gaussian and RobustSplat across all metrics.

\section{Method}
\label{sec:method}
\begin{figure*}[t]
\centering
\includegraphics[width=\textwidth]{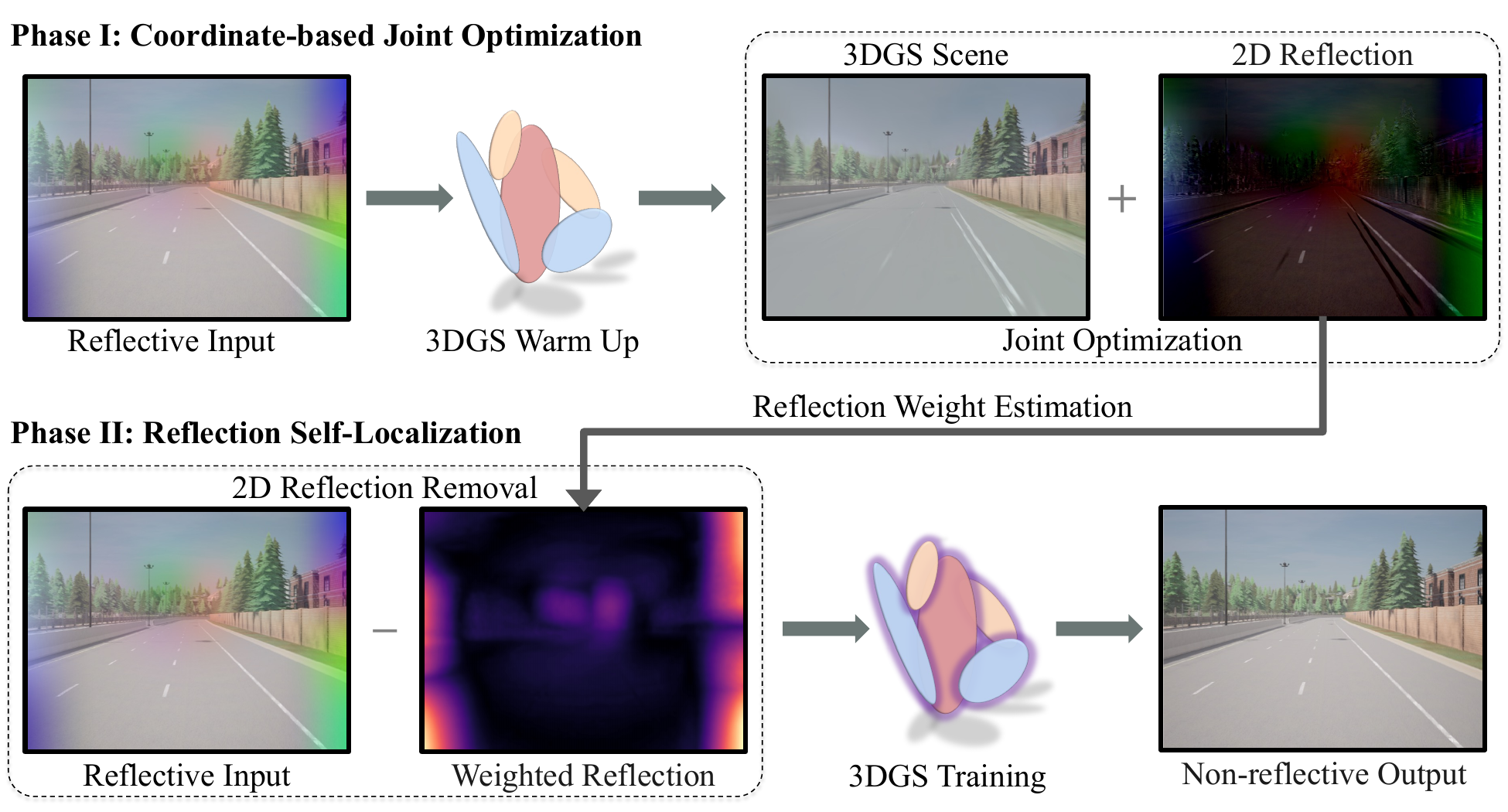}
\caption{\textbf{Overview of RIDE-Splat.} Given posed driving frames with windshield reflections, RIDE-Splat reconstructs a reflection-free 3D Gaussian Splatting (3DGS) scene without paired clean images or reflection masks. \textbf{Phase I: Coordinate-Based Joint Optimization} (Sec.~\ref{sec:joint_optimization}) warms up scene geometry, then jointly separates a world-consistent 3DGS scene from a shared, signed 2D reflection layer. \textbf{Phase II: Reflection Self-Localization} (Sec.~\ref{sec:self_local}) derives a continuous spatial weight from the recovered reflection layer and retrains the decomposition from scratch with this support fixed, preventing weak scene residuals from entering the final reconstruction. This refit adds no optimized parameters. At inference, the reflection layer is discarded and standard 3DGS rasterization renders reflection-free novel views.}
\label{fig:pipeline}
\end{figure*}

Given $N$ reflective frames $\{I_i\}_{i=1}^{N}$ and corresponding camera poses $\{\pi_i\}_{i=1}^{N}$, RIDE-Splat reconstructs a 3D Gaussian Splatting (3DGS) representation~\cite{3dgs} of the transmitted scene for reflection-free novel-view rendering, using no paired clean images or reflection masks.
Figure~\ref{fig:pipeline} summarizes the two-stage pipeline.
First, we jointly optimize the 3DGS scene $G$ and a single shared, signed reflection image $R$.
We then derive a fixed spatial weight $W$ from the recovered $R$ and refit the model from scratch.
At inference, $R$ is discarded and only $G$ is rendered.

\subsection{3D Scene and Signed Reflection Model}
Let $G=\{g_k\}_{k=1}^{K}$ denote a 3DGS scene with $K$ Gaussians.
For camera $\pi_i$, rasterization produces the transmission image $T_i=\operatorname{render}(G;\pi_i)$.

To motivate a signed reflection image, consider windshield image formation. At each pixel $u$, $\alpha(u)\in[0,1]$ is an attenuation coefficient broadcast across RGB channels, and $L_r(u)$ is the added RGB reflected radiance. The observed image is approximated by
\begin{equation}
\label{eq:physical}
I_i \approx (1-\alpha)\odot T_i+L_r = T_i+\bigl(L_r-\alpha\odot T_i\bigr).
\end{equation}
Although $L_r$ is nonnegative, the residual $L_r-\alpha\odot T_i$ is generally signed and varies with $T_i$, capturing both added reflected radiance and attenuated transmission. RIDE-Splat approximates its dominant camera-static component with a single directly optimized, signed RGB image $R$ shared across all views. The predicted observed frame is
\begin{equation}
\label{eq:obs}
\hat I_i=T_i+R\approx I_i,\qquad i=1,\ldots,N.
\end{equation}
Because $R$ is shared across views, it can represent camera-attached reflection content but cannot follow scene parallax. The reflection model requires no view conditioning, network, hash grid, or per-frame coefficients.

\subsection{Coordinate-Based Joint Optimization}
\label{sec:joint_optimization}
For a world point $X$, its pixel $p_i(X)=\pi_i(X)$ changes with camera motion, but the same 3D Gaussian can approximate its appearance in every view.
For a rigidly mounted camera, the dominant cabin reflection exhibits little image-space motion.
Encoding this reflection in the 3DGS scene $G$ would require inconsistent world geometry.
Conversely, encoding parallax-bearing road texture in the fixed image $R$ would incur error across views.
Joint optimization therefore favors scene content in $G$ and camera-static residuals in $R$.
The separation is not unique because $R$ can also absorb camera-static scene content such as the vehicle hood or a co-moving vehicle.

At iteration $t$, a geometry warm-up gates the shared image, giving the prediction
\begin{equation}
\label{eq:warmup}
\hat I_i^{(t)}=T_i+\mathbbm{1}[t>t_0]R,
\end{equation}
where $\mathbbm{1}[t>t_0]$ is an indicator function that starts optimization of $R$ after iteration $t_0$. We augment the standard 3DGS photometric loss with an $\ell_1$ penalty on $R$, weighted by $\beta$:
\begin{equation}
\label{eq:obj}
\mathcal{L}=\mathcal{L}_{\text{photo}}(\hat I_i^{(t)},I_i)+\beta\lVert R\rVert_1.
\end{equation}
Averaging $\lVert\cdot\rVert_1$ over channels and pixels keeps the reflection penalty on the same scale as the photometric objective.
The regularizer resolves ambiguous pixels in favor of the 3DGS scene, while the warm-up establishes initial scene geometry before $R$ can absorb persistent image-space residuals.

\begin{figure*}[ht]
\centering
\includegraphics[width=\textwidth]{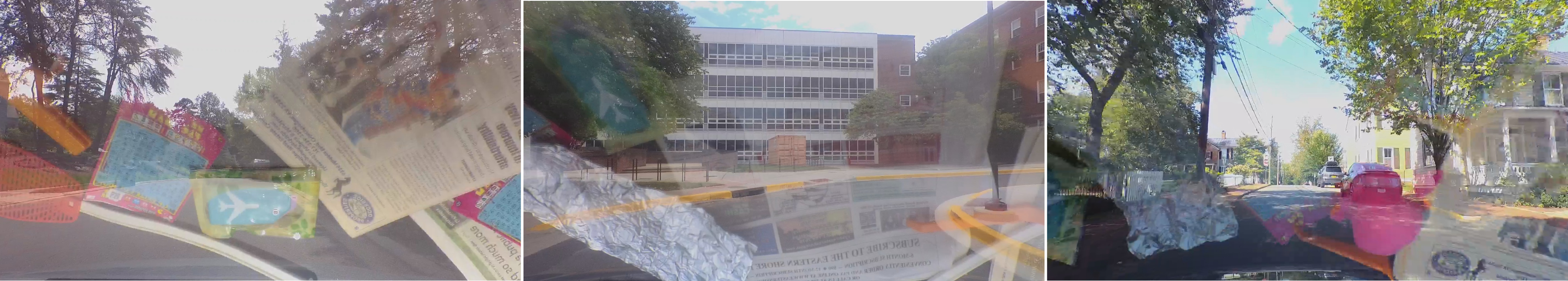}
\caption{\textbf{Reflective training views from the \emph{Leesburg}, \emph{Martin}, and \emph{Reston} scenes in the RIDE-Splat Dataset, shown from left to right.} Physical windshield reflections are produced by common cabin objects, including newspapers, aluminum foil, and clips.}
\label{fig:dataset_example}
\end{figure*}

\subsection{Reflection Self-Localization}
\label{sec:self_local}
Although sharing and sparsity constrain $R$, coordinate-based joint optimization can still leave low-amplitude scene residuals in $R$ outside reflection regions. To prevent this leakage in the final decomposition, reflection self-localization estimates a continuous spatial weight $W$ from the recovered $R$, restricting where the reflection model may act.
For each pixel $u$, we define the channel-max response $a(u)=\max_{q\in\{1,2,3\}}|R_q(u)|$. The absolute value measures reflection magnitude regardless of sign, while the channel maximum retains any strong RGB response without dilution by weaker channels.
We smooth $a$ with a Gaussian kernel $\kappa_\sigma$ of standard deviation $\sigma$ to aggregate pixel-level responses into spatially coherent support, then normalize over all pixel locations $v$:
\begin{equation}
\label{eq:weight}
\begin{aligned}
b(u) &= (\kappa_\sigma\ast a)(u), \\
W(u) &= \frac{b(u)-\min_v b(v)}{\max_v b(v)-\min_v b(v)+\epsilon}.
\end{aligned}
\end{equation}
Here $\epsilon>0$ ensures numerical stability, the normalization yields a continuous spatial weight $W(u){\in}[0,1]$, and the bandwidth $\sigma$ controls the extent of the estimated support. The scalar map $W$ determines where the signed RGB appearance $R$ can contribute.
Because $W$ is inferred only after joint optimization, it is unavailable at initialization. We therefore restart the joint optimization of $G$ and $R$ from scratch with $W$ frozen, retaining the objective in Eq.~\eqref{eq:obj} and the gated 3DGS warm-up. This trains the final decomposition under localized reflection support from the outset rather than inheriting leakage from the joint fit.
Under this weighting, Eq.~\eqref{eq:warmup} becomes
\begin{equation}
\label{eq:selfloc}
R_{\mathrm{eff}}=R\odot W,\qquad \hat I_i^{(t)}=T_i+\mathbbm{1}[t>t_0]R_{\mathrm{eff}}.
\end{equation}
The photometric term uses the weighted reflection, while the $\ell_1$ penalty remains on $R$.
In regions with little reflection evidence, $W$ suppresses the effective reflection, forcing those pixels to be explained by $G$. The component ablation in Sec.~\ref{sec:component_ablation} confirms that this self-localized refit provides additional gains over the shared reflection layer alone.
Because $W$ is frozen, the second pass introduces no additional optimized parameters.
At inference, we omit $R$ and render reflection-free novel views from $G$ using the standard 3DGS rendering path.
We ablate our choice of $\beta$ and $\sigma$ in Sec.~\ref{sec:hyperparameter_sensitivity}.

\section{RIDE-Splat Dataset}
\label{sec:ride_dataset}

Existing driving-scene 3D reconstruction benchmarks either retain exact clean targets by adding synthetic reflections or capture physical reflections without non-reflective targets. Neither setting directly measures transmission recovery under physical windshield reflections. To fill this gap, we introduce the \textbf{RIDE-Splat Dataset}, the first 3D reconstruction benchmark designed specifically for reflection removal from driving video. It pairs reflective training traversals of real driving scenes with separately captured non-reflective evaluation traversals, enabling quantitative evaluation against actual non-reflective images. Table~\ref{tab:dataset_comparison} compares the evaluation datasets.

We recorded H.264 clips in clear daytime conditions using a Garmin S35 dashcam at $1920{\times}1080$ and 29.97 FPS. To diversify reflection appearance, we placed common cabin objects made of paper, laminated paper, aluminum, plastic, painted wood, and steel on the dashboard. The object set differs by scene but remains fixed within each reflective traversal, producing scene-specific, approximately camera-static reflections. Figure~\ref{fig:dataset_example} shows representative reflective training views from all three scenes. To prepare the footage for 3D reconstruction, we removed the 40-pixel-tall dashcam metadata banner, undistorted and cropped the frames to $1900{\times}1025$, and downsampled them to reduce dataset size. Table~\ref{tab:ride_splits} reports the downsampled frame rate and number of training and evaluation frames for each scene.

\begin{table}[t]
\centering
\caption{\textbf{Evaluation datasets and RIDE-Splat splits.}}
\label{tab:datasets}
\subcaptionbox{\textbf{Dataset comparison.} XLD~\cite{xld} and Wayve~\cite{wayve} preserve exact targets by adding synthetic reflections. RIDE-Splat instead trains on physical windshield reflections and evaluates against a separately captured non-reflective traversal. Alignment between traversals is approximate.\label{tab:dataset_comparison}}[\columnwidth]{%
\setlength{\tabcolsep}{3.5pt}
\begin{tabular}{lccc}
\toprule
Dataset & Data & Reflections & Reference \tabularnewline
\midrule
XLD~\cite{xld} & Synthetic & Synthetic & Exact \tabularnewline
Wayve~\cite{wayve} & Real & Synthetic & Exact \tabularnewline
\textbf{RIDE-Splat Dataset (Ours)} & \textbf{Real} & \textbf{Real} & Approximate \tabularnewline
\bottomrule
\end{tabular}%
}
\par\medskip
\subcaptionbox{\textbf{RIDE-Splat Dataset scene splits.} The footage is downsampled from 29.97 FPS to the listed rates. Reflective views are used only for training, and uniformly sampled non-reflective views are held out for evaluation. Additional captured scenes will be publicly released.\label{tab:ride_splits}}[\columnwidth]{%
\setlength{\tabcolsep}{3pt}
\begin{tabular}{lccc}
\toprule
& & \multicolumn{2}{c}{Number of Frames} \tabularnewline
\cmidrule(lr){3-4}
Scene & FPS & Reflective Training & Non-Reflective Evaluation \tabularnewline
\midrule
\emph{Leesburg} & 8 & 134 & 30 \tabularnewline
\emph{Martin} & 6 & 82 & 30 \tabularnewline
\emph{Reston} & 8 & 52 & 30 \tabularnewline
\midrule
\textbf{Total} & -- & \textbf{268} & \textbf{90} \tabularnewline
\bottomrule
\end{tabular}%
}
\end{table}

The evaluated benchmark contains three scenes, \emph{Leesburg}, \emph{Martin}, and \emph{Reston}, spanning a historic town center, a tree-lined institutional district, and suburban residential streets. These settings vary in traffic, building density, tree cover, and lateral parallax. For each scene, we recorded a reflective training traversal and a non-reflective reference traversal of the same route within minutes of one another. The camera remained in the same rigid mount, and we removed only the reflection-inducing materials between traversals. To reduce cross-traversal confounds, we visually selected sequences with few transient objects and minimal visible change. We jointly registered both traversals with COLMAP~\cite{colmap}, placing their camera poses in a common coordinate frame. All geometry and camera poses were recovered from RGB footage alone, without LiDAR, odometry, depth, or other vehicle sensors.

Only reflective frames are used for training, without reflection masks or annotations, so methods must infer contaminated regions directly. Evaluation frames are uniformly sampled from each non-reflective traversal. Each method renders at these held-out camera poses and is scored against the corresponding non-reflective images.

The fixed camera mount, short capture interval, and sequence selection reduce cross-traversal variation, but traffic, illumination, and camera trajectories cannot be matched exactly. The resulting references are therefore approximate. Nevertheless, they contain minimal windshield reflections and are never used during optimization, so they directly evaluate transmission recovery under physical reflections. We will publicly release the RIDE-Splat Dataset, including the evaluated benchmark and additional captured scenes.

\begin{table*}[t]
\centering
\caption{\textbf{Quantitative results on the RIDE-Splat Dataset, XLD~\cite{xld}, and Wayve~\cite{wayve}.} Metrics are averaged across the evaluated scenes in each dataset, and the best result is bolded. Because DC-Gaussian exhibits substantial underexposure and color shifts, we also report a \emph{+ Oracle GT Color Correction} variant that applies the released color transform using each test target as reference. This variant deliberately favors DC-Gaussian and is excluded from bolding. Even so, RIDE-Splat achieves the strongest overall results.}
\label{tab:main_comparison}
\begin{tabular}{lccccccccc}
\toprule
& \multicolumn{3}{c}{RIDE-Splat Dataset (Ours)} & \multicolumn{3}{c}{XLD~\cite{xld}} & \multicolumn{3}{c}{Wayve~\cite{wayve}} \\
\cmidrule(lr){2-4} \cmidrule(lr){5-7} \cmidrule(lr){8-10}
Method & PSNR~$\uparrow$ & SSIM~$\uparrow$ & LPIPS~$\downarrow$
       & PSNR~$\uparrow$ & SSIM~$\uparrow$ & LPIPS~$\downarrow$
       & PSNR~$\uparrow$ & SSIM~$\uparrow$ & LPIPS~$\downarrow$ \\
\midrule
3DGS~\cite{3dgs} & 17.37 & 0.7204 & 0.5445 & 21.37 & 0.8824 & 0.4882 & 20.14 & 0.7366 & 0.4426 \\
DC-Gaussian~\cite{dcgaussian} & 14.47 & 0.6821 & 0.5549 & 18.32 & 0.8436 & 0.5276 & 20.07 & 0.7161 & 0.4694 \\
\textit{+ Oracle GT Color Correction} & \textit{20.25} & \textit{0.7442} & \textit{0.5109} & \textit{23.07} & \textit{0.8894} & \textit{0.4920} & \textit{20.49} & \textit{0.7155} & \textit{0.4713} \\
RobustSplat~\cite{robustsplat} & 18.30 & 0.7317 & 0.5216 & 21.64 & 0.8815 & 0.4891 & 20.19 & 0.7364 & 0.4424 \\
\textbf{RIDE-Splat (Ours)} & \textbf{18.48} & \textbf{0.7428} & \textbf{0.4873} & \textbf{25.23} & \textbf{0.8980} & \textbf{0.4006} & \textbf{20.84} & \textbf{0.7422} & \textbf{0.4179} \\
\bottomrule
\end{tabular}
\end{table*}

\section{Experiments}
\label{sec:exp}

We evaluate RIDE-Splat on fully synthetic scenes, real driving imagery with synthetic reflections, and real driving imagery with physical windshield reflections. Together, these settings test whether it consistently recovers the transmitted scene as both scene content and reflection formation become more realistic.
All methods receive identical reflective training images and camera poses, while non-reflective targets are used only for evaluation.
Beyond the main comparison, we isolate the contributions of the Reflection Layer and self-localization, evaluate hyperparameter sensitivity, and stress-test the camera-static reflection prior under increasing drift.

\subsection{Datasets}
\label{sec:datasets}
We evaluate on three datasets that vary both scene content and reflection formation.
The RIDE-Splat Dataset pairs real driving scenes with physical windshield reflections. We use the \emph{Leesburg}, \emph{Martin}, and \emph{Reston} splits in Table~\ref{tab:datasets}.
XLD~\cite{xld} pairs synthetic scenes with synthetic reflections, while Wayve~\cite{wayve} combines real driving imagery with synthetic reflections.

For XLD, we evaluate all six scenes: \emph{Town01}, \emph{Town02}, \emph{Town03}, \emph{Town04}, \emph{Town05}, and \emph{Town10}.
We add a camera-static synthetic windshield reflection to 150 training images per town and evaluate all 28 clean views captured at a 1\,m left lateral offset.
For Wayve, we use the first five 10-second sequences, \emph{scene\_001--scene\_005}, and add static synthetic reflections to 176 training images per sequence.
Every eighth original frame is held out, yielding 24 clean evaluation images per sequence.
Appendix Sec.~\ref{sec:synthetic_reflection_parameters} details the synthetic-reflection construction and parameters for both datasets.

\subsection{Baselines and Metrics}
We use the official implementations and published settings for all baselines. Every method receives identical training images and camera poses and is evaluated against identical targets.
We compare against 3DGS~\cite{3dgs}, DC-Gaussian~\cite{dcgaussian}, and RobustSplat~\cite{robustsplat}.
DC-Gaussian specifically targets driving video with windshield reflections, while RobustSplat is the strongest single-model anti-distractor method in DF3DV-1K~\cite{lu2026df3dv}.
We report full-image PSNR, SSIM, and LPIPS.

\subsection{DC-Gaussian Evaluation Protocol}
\label{sec:dc_evaluation}
DC-Gaussian~\cite{dcgaussian} introduces DCVR, a self-captured real-world dataset with strong reflections, and also evaluates on BDD100K~\cite{bdd100k}.
Both are real-world dashcam datasets with windshield obstructions, but neither provides non-reflective transmission targets.
DC-Gaussian decomposes each view into a transmission image and an image-space obstruction image, then combines them using a learned opacity map.
Its published PSNR, SSIM, and LPIPS compare this recomposed output with held-out observed frames.
These metrics therefore measure whether the two predicted layers reproduce the observed obstructed frame rather than whether the transmission recovers the reflection-free scene.
Our metrics instead compare transmission renders directly with non-reflective targets: exact clean images for XLD and Wayve, and separately captured images for the RIDE-Splat Dataset.

On real-world data, DC-Gaussian transmission renders exhibit substantial underexposure and color shifts.
We report the unmodified renders as DC-Gaussian.
To prevent these errors from dominating the comparison, the \emph{+ Oracle GT Color Correction} variant uses the released routine to fit one global quadratic color transform from each transmission render to its test target over all unsaturated pixel correspondences.
This transform is not part of the published method or default evaluation and could be applied to any renderer.
Because it uses test-target oracle information, the variant deliberately favors DC-Gaussian and is excluded from best-result highlighting.

\subsection{Implementation Details}
RIDE-Splat uses the original 3DGS optimization settings for $G$.
We optimize $R$ with Adam at a learning rate $\eta_R=0.01$ and set $\beta=0.01$.
Following RobustSplat~\cite{robustsplat}, we delay activation of the reflection image behind a geometry warm-up, ending at $t_0=2000$ iterations. This places activation after densification begins at iteration 500, and well before it concludes, thus scene geometry is established before $R$ can compete for it.
The self-localization pass uses $\sigma=8$ and repeats the same schedule with $W$ fixed.
We ablate $\eta_R$, $\beta$, and $\sigma$ in Sec.~\ref{sec:hyperparameter_sensitivity}.
No pretrained image generator or reflection-removal network is used.

\section{Results}
\label{sec:results}

\begin{figure*}[t]
\centering
\includegraphics[width=\textwidth]{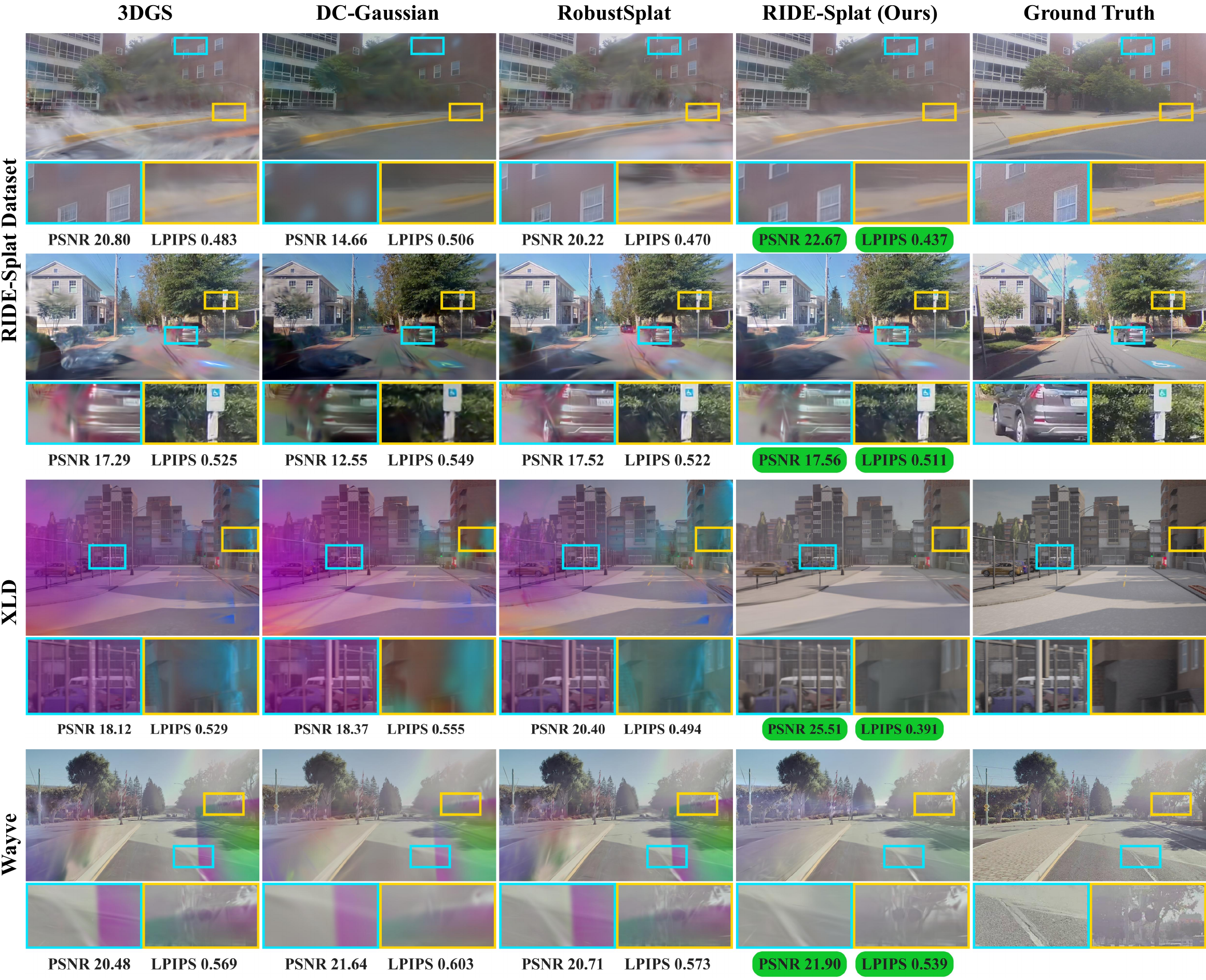}
\caption{\textbf{Qualitative results on the RIDE-Splat Dataset, XLD~\cite{xld}, and Wayve~\cite{wayve}.} The examples show \emph{Martin} and \emph{Leesburg} from the RIDE-Splat Dataset, \emph{Town02} from XLD, and \emph{scene\_003} from Wayve. Compared with 3DGS~\cite{3dgs}, DC-Gaussian~\cite{dcgaussian}, and RobustSplat~\cite{robustsplat}, RIDE-Splat more effectively removes windshield reflections while preserving the transmitted scene. DC-Gaussian exhibits visible color shifts, so Table~\ref{tab:main_comparison} also reports the \emph{+ Oracle GT Color Correction} variant described in Section~\ref{sec:dc_evaluation} that uses each test target as reference.}
\label{fig:qualitative}
\end{figure*}

Table~\ref{tab:main_comparison} evaluates transmission quality against held-out non-reflective targets. The RIDE-Splat Dataset uses separately captured, approximately aligned non-reflective traversals, whereas XLD and Wayve use the original clean frames before reflections are added to the training images. Except for the explicitly marked color-corrected variant, test targets are used only for evaluation. Among methods without test-target access, RIDE-Splat ranks first on all nine aggregate metrics across physical and synthetically added reflections. Its largest margins occur on XLD, where it improves PSNR by 3.59\,dB and reduces LPIPS by 0.0876 relative to the next-best baseline. Appendix Sec.~\ref{sec:per_scene_metrics} reports per-scene results. Excluding the target-aware oracle correction, RIDE-Splat outperforms the baselines on 38 of 42 scene-metric comparisons, confirming that the aggregate gains are not driven by a small subset of scenes.

Because DC-Gaussian exhibits substantial underexposure and color shifts, Table~\ref{tab:main_comparison} also reports the \emph{+ Oracle GT Color Correction} variant described in Sec.~\ref{sec:dc_evaluation}. Even with this test-target access, DC-Gaussian remains behind RIDE-Splat on every metric for XLD and Wayve. On the RIDE-Splat Dataset, the corrected variant leads in PSNR and SSIM, while RIDE-Splat retains the best LPIPS. These target-aware scores are excluded from best-result highlighting.

Figure~\ref{fig:qualitative} presents representative comparisons from all three datasets. Vanilla 3DGS embeds windshield reflections into the reconstruction, while DC-Gaussian leaves residual reflections and visible color shifts. RobustSplat introduces broader image artifacts on real scenes. RIDE-Splat more effectively removes the reflections while preserving the transmitted road scene, yielding sharper building and road-sign textures than DC-Gaussian and fewer artifacts than RobustSplat. These visual gains confirm that the metric improvements reflect cleaner transmission rather than simply a global appearance change.

\subsection{Component Ablation}
\label{sec:component_ablation}
Table~\ref{tab:component_ablation} applies the method components cumulatively across XLD. Our coordinate-based joint-optimization method, described in Sec.~\ref{sec:joint_optimization}, provides the primary improvement across all metrics. Reflection self-localization, described in Sec.~\ref{sec:self_local}, further improves all metrics without adding optimized parameters. The ablation confirms that coordinate-based separation drives the gain and that self-localization further refines the reconstructed scene.

\subsection{Hyperparameter Sensitivity}
\label{sec:hyperparameter_sensitivity}
We vary one hyperparameter at a time around the defaults $\eta_R=0.01$, $\beta=0.01$, and $\sigma=8$. Table~\ref{tab:hyperparameter_sensitivity} reports the results on \emph{Town01}. Performance remains stable across the tested learning rates. Reducing $\beta$ to 0.002 increases the mean recovered $\lvert R\rvert$ by 40\% while leaving PSNR and SSIM nearly unchanged, suggesting that sharing the reflection image across views, rather than sparsity alone, prevents it from absorbing scene content. Increasing $\beta$ to 0.05 over-regularizes the reflection image and degrades all metrics. Among the tested smoothing scales, $\sigma=32$ yields the best LPIPS, whereas the narrower support at $\sigma=2$ lowers PSNR.

\subsection{Robustness to Reflection Motion}
\label{sec:motion_robustness}
We stress-test the camera-static assumption on \emph{Town01} with a single seed using the Reflection Layer from the \textit{Coordinate-Based Joint Optimization} row from Table~\ref{tab:component_ablation}. Within a fixed support, the reflection appearance $L_r$ is translated horizontally by the specified number of pixels per frame over 150 frames. At the largest drift of 1.2 pixels per frame, the cumulative displacement reaches 180 pixels, or 9.4\% of the image width. Across the sweep, Reflection Layer PSNR varies by only 0.65\,dB, compared with 1.45\,dB for DC-Gaussian. The shared reflection model therefore tolerates moderate horizontal drift, although reflections with independent parallax or deformation remain outside its scope.

\begin{table}[t]
\centering\small
\caption{\textbf{Component ablation on XLD~\cite{xld}.} Coordinate-based Joint Optimization (Section~\ref{sec:joint_optimization}) and Reflection Self-Localization (Section~\ref{sec:self_local}) are applied cumulatively, metrics are averaged across all six towns, and the best value in each column is bolded.}
\label{tab:component_ablation}
\begin{tabular}{lccc}
\toprule
Configuration & PSNR~$\uparrow$ & SSIM~$\uparrow$ & LPIPS~$\downarrow$ \\
\midrule
3DGS~\cite{3dgs} & 21.37 & 0.8824 & 0.4882 \\
+ Joint Optimization & 24.78 & 0.8968 & \textbf{0.3994} \\
+ Self-Localization & \textbf{25.23} & \textbf{0.8980} & 0.4006 \\
\bottomrule
\end{tabular}
\end{table}

\begin{table}[t]
\centering\small
\caption{\textbf{Hyperparameter sensitivity on \emph{Town01} from XLD~\cite{xld}.} Each row varies one default value with the others held fixed. The default row is the run reported in Table~\ref{tab:main_comparison}.}
\label{tab:hyperparameter_sensitivity}
\begin{tabular}{cccccc}
\toprule
$\eta_R$ & $\beta$ & $\sigma$ & PSNR~$\uparrow$ & SSIM~$\uparrow$ & LPIPS~$\downarrow$ \\
\midrule
0.01 & 0.01 & 8 & 25.25 & 0.9255 & 0.4586 \\
\midrule
0.005 & 0.01 & 8 & 25.10 & 0.9264 & 0.4572 \\
0.05 & 0.01 & 8 & 25.34 & 0.9263 & 0.4640 \\
\midrule
0.01 & 0.002 & 8 & 25.24 & 0.9269 & 0.4728 \\
0.01 & 0.05 & 8 & 22.69 & 0.9197 & 0.5045 \\
\midrule
0.01 & 0.01 & 2 & 24.82 & 0.9235 & 0.4590 \\
0.01 & 0.01 & 32 & 25.32 & 0.9249 & 0.4444 \\
\bottomrule
\end{tabular}
\end{table}

\begin{table}[t]
\centering\small
\caption{\textbf{PSNR on \emph{Town01} from XLD~\cite{xld} under increasing reflection drift.} The reflection appearance is translated horizontally within a fixed support by the stated number of pixels per frame over 150 frames. Range is the difference between the minimum and maximum PSNR across the tested drift levels.}
\label{tab:motion_robustness}
\begin{tabular}{ccc}
\toprule
 & \multicolumn{2}{c}{PSNR~$\uparrow$} \tabularnewline
\cmidrule(lr){2-3}
Drift (Pixels) & Reflection Layer (Ours) & DC-Gaussian \tabularnewline
\midrule
0.0 & 24.78 & 12.97 \tabularnewline
0.15 & 24.75 & 12.04 \tabularnewline
0.3 & 24.62 & 13.02 \tabularnewline
0.5 & 24.84 & 12.21 \tabularnewline
0.8 & 24.46 & 13.11 \tabularnewline
1.2 & 24.19 & 13.49 \tabularnewline
\midrule
Range & 0.65 & 1.45 \tabularnewline
\bottomrule
\end{tabular}
\vspace{-5mm}
\end{table}

\section{Limitations}
RIDE-Splat assumes that the dominant windshield reflection is semi-transparent and approximately fixed in image coordinates. The motion stress test indicates tolerance to moderate drift, but substantial independent motion or deformation may leave residual contamination. As a 3DGS-based method, it also inherits challenges under sparse forward-facing views, although its reflection formulation is orthogonal to driving-scene extensions such as Street Gaussians~\cite{streetgaussians} and can be combined with them for large, dynamic environments. Evaluating real windshield reflections against non-reflective targets requires separate traversals, making exact temporal and environmental matching impossible and leaving the references approximate. Our physical-reflection evaluation currently covers a three-scene subset of the broader RIDE-Splat Dataset, and additional captured scenes will be publicly released.

\section{Applications and Future Work}
RIDE-Splat can convert existing dashcam video into reflection-free 3D assets for closed-loop simulation and rare-event replay. Beyond driving, its camera-coordinate prior may extend to robot cameras behind protective windows or transparent housings, where enclosure reflections move with the camera rather than the environment. Future work can relax the shared-image assumption through constrained low-dimensional reflection motion and adapt the self-localization weight using multi-view consistency, without introducing unrestricted per-view capacity that can absorb scene content. Expanding the RIDE-Splat Dataset across weather, lighting, camera systems, and reflection dynamics, together with more synchronized reference capture, will broaden evaluation and improve reference fidelity.

\section{Conclusion}
RIDE-Splat reframes windshield reflection removal as a coordinate-separation problem for 3D reconstruction from posed driving video. It assigns world-consistent transmission through the windshield to a 3D Gaussian Splatting (3DGS)~\cite{3dgs} scene and the dominant camera-attached reflection to a shared, signed image, using camera motion to separate them without clean supervision. Reflection self-localization further restricts where the reflection layer can act in a second fit without adding optimized parameters, and discarding the layer at inference preserves the standard 3DGS rendering path. The RIDE-Splat Dataset provides the first 3D reconstruction benchmark for reflection removal from driving video, pairing training traversals containing physical windshield reflections with held-out non-reflective captures. Across the RIDE-Splat Dataset, XLD~\cite{xld}, and Wayve~\cite{wayve}, RIDE-Splat outperforms 3DGS, DC-Gaussian~\cite{dcgaussian}, and RobustSplat~\cite{robustsplat} on every aggregate metric; per-scene and qualitative comparisons confirm that this advantage reflects cleaner transmission rather than a global appearance correction. Component ablations identify the shared reflection image as the primary source of improvement and self-localization as a consistent refinement. More broadly, modeling structured contamination in the coordinate system where it remains stable provides a simple principle for reconstructing scenes from driving archives and other video captured through glass.

{\small
\bibliographystyle{IEEEtran}
\bibliography{references}
}
\onecolumn
\section*{Acknowledgments}
\label{sec:acknowledgments}
This research is based upon work supported by the Office of the Director of National Intelligence (ODNI), Intelligence Advanced Research Projects Activity (IARPA), via IARPA R\&D Contract No. 140D0423C0076. The views and conclusions contained herein are those of the authors and should not be interpreted as necessarily representing the official policies or endorsements, either expressed or implied, of the ODNI, IARPA, or the U.S. Government. The U.S. Government is authorized to reproduce and distribute reprints for Governmental purposes notwithstanding any copyright annotation thereon. Commercial support was provided by the Amazon Research Awards program and Open Philanthropy. Further support was provided by the NSF TRAILS Institute (2229885).

\appendix
\subsection{Synthetic Reflection Construction}
\label{sec:synthetic_reflection_parameters}
Table~\ref{tab:synthetic_reflection_parameters} reports the synthesis parameters for XLD and Wayve.
Both settings use static reflections with zero flicker and image-space drift.
For Wayve, reflection support $\alpha$ is derived from a real windshield annotation, covers 31.8\% of the image, and is shared across \emph{scene\_001--scene\_005}.
The support is used only to synthesize reflective training images and is never provided to a method.
The added reflection $L_r$ is a synthetic low-frequency wash with strength $0.5$--$0.7$ for XLD and $0.5$ for Wayve. Wayve also uses no center glare and lower obstruction strength, producing milder contamination that may contribute to the smaller PSNR gap between methods.

\begin{table}[h]
\centering\small
\caption{\textbf{Synthetic reflection parameters.} XLD~\cite{xld} and Wayve~\cite{wayve} use the same compositing pipeline without temporal variation. A dash indicates that no separate value is specified.}
\label{tab:synthetic_reflection_parameters}
\begin{tabular}{lcc}
\toprule
Parameter & XLD & Wayve \tabularnewline
\midrule
Flicker & 0.0 & 0.0 \tabularnewline
Drift [px] & 0 & 0 \tabularnewline
Center glare & 0.35 & 0.0 \tabularnewline
Feather & 8.0 & 6.0 \tabularnewline
Obstruction blur & 14.0 & 12.0 \tabularnewline
Obstruction strength & 0.6 & 0.4 \tabularnewline
Side fraction & 0.22 & 0.0 \tabularnewline
$L_r$ strength & 0.5–0.7 & 0.5 \tabularnewline
\bottomrule
\end{tabular}
\end{table}

\subsection{Per-Scene Metrics}
\label{sec:per_scene_metrics}
Tables~\ref{tab:per_scene_psnr}, \ref{tab:per_scene_ssim}, and \ref{tab:per_scene_lpips} report scene-level performance for the three RIDE-Splat Dataset scenes, all six XLD towns, and five Wayve sequences.
They use the same methods and evaluation protocols as the main comparison in Table~\ref{tab:main_comparison}.
Excluding the test-target correction, RIDE-Splat achieves the best PSNR on 12 of 14 scenes and the best SSIM and LPIPS on 13 of 14, leading 38 of 42 scene-metric comparisons. RobustSplat leads PSNR on \emph{Reston}, DC-Gaussian leads PSNR on \emph{scene\_002}, and 3DGS leads SSIM and LPIPS on \emph{scene\_001}.

\begin{table*}[t]
\centering
\caption{\textbf{PSNR on each scene in the RIDE-Splat Dataset, XLD~\cite{xld}, and Wayve~\cite{wayve}.} The best result for each scene is bolded. The DC-Gaussian \emph{+ Oracle GT Color Correction} variant is excluded from bolding because it uses test-target information.}
\label{tab:per_scene_psnr}
\resizebox{\textwidth}{!}{%
\begin{tabular}{l*{17}{c}}
\toprule
& \multicolumn{4}{c}{RIDE-Splat Dataset (Ours)} & \multicolumn{7}{c}{XLD~\cite{xld}} & \multicolumn{6}{c}{Wayve~\cite{wayve}} \tabularnewline
\cmidrule(lr){2-5} \cmidrule(lr){6-12} \cmidrule(lr){13-18}
Method & \emph{Leesburg} & \emph{Martin} & \emph{Reston} & Mean & \emph{Town01} & \emph{Town02} & \emph{Town03} & \emph{Town04} & \emph{Town05} & \emph{Town10} & Mean & \emph{s001} & \emph{s002} & \emph{s003} & \emph{s004} & \emph{s005} & Mean \tabularnewline
\midrule
3DGS~\cite{3dgs} & 15.55 & 18.64 & 17.91 & 17.37 & 20.97 & 20.38 & 22.13 & 21.63 & 22.07 & 21.03 & 21.37 & 19.97 & 19.84 & 20.16 & 20.58 & 20.16 & 20.14 \tabularnewline
RobustSplat~\cite{robustsplat} & 15.37 & 20.10 & \textbf{19.42} & 18.30 & 21.22 & 20.76 & 22.21 & 21.91 & 22.52 & 21.22 & 21.64 & 19.86 & 20.23 & 20.18 & 20.57 & 20.10 & 20.19 \tabularnewline
DC-Gaussian~\cite{dcgaussian} & 13.73 & 15.67 & 14.01 & 14.47 & 12.36 & 20.43 & 19.47 & 15.78 & 20.22 & 21.63 & 18.32 & 19.87 & \textbf{20.64} & 19.38 & 20.39 & 20.07 & 20.07 \tabularnewline
\textit{+ Oracle GT Color Correction} & \textit{17.10} & \textit{22.84} & \textit{20.80} & \textit{20.25} & \textit{22.69} & \textit{21.15} & \textit{23.56} & \textit{24.48} & \textit{24.00} & \textit{22.55} & \textit{23.07} & \textit{20.12} & \textit{20.70} & \textit{20.99} & \textit{20.52} & \textit{20.11} & \textit{20.49} \tabularnewline
\textbf{RIDE-Splat (Ours)} & \textbf{16.03} & \textbf{20.18} & 19.21 & \textbf{18.48} & \textbf{25.25} & \textbf{24.72} & \textbf{26.00} & \textbf{26.07} & \textbf{25.31} & \textbf{24.02} & \textbf{25.23} & \textbf{20.10} & 20.54 & \textbf{20.66} & \textbf{21.74} & \textbf{21.15} & \textbf{20.84} \tabularnewline
\bottomrule
\end{tabular}%
}
\end{table*}

\begin{table*}[t]
\centering
\caption{\textbf{SSIM on each scene in the RIDE-Splat Dataset, XLD~\cite{xld}, and Wayve~\cite{wayve}.} The best result for each scene is bolded. The DC-Gaussian \emph{+ Oracle GT Color Correction} variant is excluded from bolding because it uses test-target information.}
\label{tab:per_scene_ssim}
\resizebox{\textwidth}{!}{%
\begin{tabular}{l*{17}{c}}
\toprule
& \multicolumn{4}{c}{RIDE-Splat Dataset (Ours)} & \multicolumn{7}{c}{XLD~\cite{xld}} & \multicolumn{6}{c}{Wayve~\cite{wayve}} \tabularnewline
\cmidrule(lr){2-5} \cmidrule(lr){6-12} \cmidrule(lr){13-18}
Method & \emph{Leesburg} & \emph{Martin} & \emph{Reston} & Mean & \emph{Town01} & \emph{Town02} & \emph{Town03} & \emph{Town04} & \emph{Town05} & \emph{Town10} & Mean & \emph{s001} & \emph{s002} & \emph{s003} & \emph{s004} & \emph{s005} & Mean \tabularnewline
\midrule
3DGS~\cite{3dgs} & 0.5940 & 0.8136 & 0.7538 & 0.7204 & 0.9077 & 0.8734 & 0.9080 & 0.9255 & 0.8917 & 0.7878 & 0.8824 & \textbf{0.7996} & 0.7325 & 0.7188 & 0.6926 & 0.7396 & 0.7366 \tabularnewline
RobustSplat~\cite{robustsplat} & 0.5940 & 0.8344 & 0.7669 & 0.7317 & 0.9102 & 0.8748 & 0.9069 & 0.9215 & 0.8920 & 0.7838 & 0.8815 & 0.7959 & 0.7375 & 0.7201 & 0.6900 & 0.7387 & 0.7364 \tabularnewline
DC-Gaussian~\cite{dcgaussian} & 0.5498 & 0.7923 & 0.7040 & 0.6821 & 0.7388 & 0.8703 & 0.8990 & 0.8711 & 0.8915 & 0.7909 & 0.8436 & 0.7790 & 0.7300 & 0.7001 & 0.6537 & 0.7180 & 0.7161 \tabularnewline
\textit{+ Oracle GT Color Correction} & \textit{0.6085} & \textit{0.8446} & \textit{0.7797} & \textit{0.7442} & \textit{0.9143} & \textit{0.8785} & \textit{0.9136} & \textit{0.9334} & \textit{0.9009} & \textit{0.7959} & \textit{0.8894} & \textit{0.7769} & \textit{0.7289} & \textit{0.7039} & \textit{0.6499} & \textit{0.7178} & \textit{0.7155} \tabularnewline
\textbf{RIDE-Splat (Ours)} & \textbf{0.6070} & \textbf{0.8420} & \textbf{0.7794} & \textbf{0.7428} & \textbf{0.9255} & \textbf{0.8931} & \textbf{0.9188} & \textbf{0.9400} & \textbf{0.9020} & \textbf{0.8084} & \textbf{0.8980} & 0.7950 & \textbf{0.7396} & \textbf{0.7251} & \textbf{0.7022} & \textbf{0.7491} & \textbf{0.7422} \tabularnewline
\bottomrule
\end{tabular}%
}
\end{table*}

\begin{table*}[t]
\centering
\caption{\textbf{LPIPS on each scene in the RIDE-Splat Dataset, XLD~\cite{xld}, and Wayve~\cite{wayve}.} The best result for each scene is bolded. The DC-Gaussian \emph{+ Oracle GT Color Correction} variant is excluded from bolding because it uses test-target information.}
\label{tab:per_scene_lpips}
\resizebox{\textwidth}{!}{%
\begin{tabular}{l*{17}{c}}
\toprule
& \multicolumn{4}{c}{RIDE-Splat Dataset (Ours)} & \multicolumn{7}{c}{XLD~\cite{xld}} & \multicolumn{6}{c}{Wayve~\cite{wayve}} \tabularnewline
\cmidrule(lr){2-5} \cmidrule(lr){6-12} \cmidrule(lr){13-18}
Method & \emph{Leesburg} & \emph{Martin} & \emph{Reston} & Mean & \emph{Town01} & \emph{Town02} & \emph{Town03} & \emph{Town04} & \emph{Town05} & \emph{Town10} & Mean & \emph{s001} & \emph{s002} & \emph{s003} & \emph{s004} & \emph{s005} & Mean \tabularnewline
\midrule
3DGS~\cite{3dgs} & 0.6057 & 0.4540 & 0.5737 & 0.5445 & 0.5341 & 0.4827 & 0.4112 & 0.5061 & 0.4588 & 0.5364 & 0.4882 & \textbf{0.3321} & 0.4665 & 0.5059 & 0.4626 & 0.4461 & 0.4426 \tabularnewline
RobustSplat~\cite{robustsplat} & 0.6007 & 0.4384 & 0.5257 & 0.5216 & 0.5296 & 0.4831 & 0.4153 & 0.5016 & 0.4567 & 0.5486 & 0.4891 & 0.3400 & 0.4575 & 0.5004 & 0.4676 & 0.4464 & 0.4424 \tabularnewline
DC-Gaussian~\cite{dcgaussian} & 0.6011 & 0.4799 & 0.5838 & 0.5549 & 0.6167 & 0.4921 & 0.4243 & 0.5418 & 0.5388 & 0.5520 & 0.5276 & 0.3689 & 0.4715 & 0.5358 & 0.4921 & 0.4787 & 0.4694 \tabularnewline
\textit{+ Oracle GT Color Correction} & \textit{0.5951} & \textit{0.4231} & \textit{0.5146} & \textit{0.5109} & \textit{0.5685} & \textit{0.4922} & \textit{0.4195} & \textit{0.4787} & \textit{0.4456} & \textit{0.5478} & \textit{0.4920} & \textit{0.3734} & \textit{0.4768} & \textit{0.5339} & \textit{0.4962} & \textit{0.4762} & \textit{0.4713} \tabularnewline
\textbf{RIDE-Splat (Ours)} & \textbf{0.5712} & \textbf{0.4025} & \textbf{0.4880} & \textbf{0.4873} & \textbf{0.4586} & \textbf{0.3776} & \textbf{0.3429} & \textbf{0.3920} & \textbf{0.3740} & \textbf{0.4584} & \textbf{0.4006} & 0.3376 & \textbf{0.4332} & \textbf{0.4700} & \textbf{0.4388} & \textbf{0.4100} & \textbf{0.4179} \tabularnewline
\bottomrule
\end{tabular}%
}
\end{table*}

\end{document}